\documentclass{article}

\usepackage[colorlinks,linkcolor=blue]{hyperref}

\usepackage[preprint]{neurips_2023}

\usepackage[utf8]{inputenc} 
\usepackage[T1]{fontenc}    
\usepackage{hyperref}       
\usepackage{url}            
\usepackage{booktabs}       
\usepackage{amsfonts}       
\usepackage{nicefrac}       
\usepackage{microtype}      
\usepackage{xcolor}         

\usepackage{amsmath,bbm,bm,graphicx,amsthm}
\usepackage{subfigure}
\usepackage{wrapfig}
\usepackage{algorithm}
\usepackage{algorithmic}
\usepackage{multicol}
\usepackage{bbding}
\usepackage{multirow}

\usepackage[utf8]{inputenc}
\usepackage{authblk}

\usepackage{pifont}

\newtheorem{define}{Definition}

\DeclareMathOperator*{\argmax}{argmax}

\title{Not All Noises Are Created Equally: \\Diffusion Noise Selection and Optimization}

\author[1]{Zipeng Qi}
\author[1]{Lichen Bai}
\author[2]{Haoyi Xiong}
\author[1,$\dagger$]{Zeke Xie}
\affil[1]{The Hong Kong University of Science and Technology (Guangzhou)}
\affil[2]{Microsoft}

\affil[$\dagger$]{Correspondence to: \textit{zekexie@hkust-gz.edu.cn}}
  
\begin{document}
\maketitle

\begin{abstract}

Diffusion models that can generate high-quality data from randomly sampled Gaussian noises have become the mainstream generative method in both academia and industry. Are randomly sampled Gaussian noises equally good for diffusion models? While a large body of works tried to understand and improve diffusion models, previous works overlooked the possibility to select or optimize the sampled noise the possibility of selecting or optimizing sampled noises for improving diffusion models. In this paper, we mainly made three contributions. First, we report that not all noises are created equally for diffusion models. We are the first to hypothesize and empirically observe that the generation quality of diffusion models significantly depend on the noise inversion stability. This naturally provides us a noise selection method according to the inversion stability. Second, we further propose a novel noise optimization method that actively enhances the inversion stability of arbitrary given noises. Our method is the first one that works on noise space to generally improve generated results without fine-tuning diffusion models. Third, our extensive experiments demonstrate that the proposed noise selection and noise optimization methods both significantly improve representative diffusion models, such as SDXL and SDXL-turbo, in terms of human preference and other objective evaluation metrics. For example, the human preference winning rates of noise selection and noise optimization over the baselines can be up to \textbf{57\%} and \textbf{72.5\%}, respectively, on DrawBench.

\end{abstract}

\section{Introduction}
\label{sec:intro}
Generative diffusion models, renowned for the impressive performance \citep{dhariwal2021diffusion}, serve as the mainstream generative paradigm with wide applications in image generation  \citep{nichol2021glide, zhang2023adding, saharia2022photorealistic}, image editing  \citep{qi2023layered, kawar2023imagic}, 3D generation  \citep{gupta20233dgen, erkocc2023hyperdiffusion}, and video generation  \citep{ho2022imagen, ho2022video}. Diffusion-based Generative AI products attracted much attention and a large number users in recent years. Understanding and improving the capabilities of diffusion models has become an essentially important topic in machine learning.


A large body of works \citep{song2020denoising, fang2023structural, podell2023sdxl, sauer2023adversarial, ho2022cascaded, lin2024sdxl} tried to enhance the generated results by working on model weight and architecture space. The importance of noise space is largely overlooked by previous studies, while it is known that diffusion models can generate diverse results, which, of course, contain good ones and bad ones. However, existing works suggest random noises are equal and fail to explore how the noises affect the quality of generated results. 

In this work, we try to visit two fundamental issues on noise space of diffusion models. First, is it possible to select better noise according to some quantitative metric? Second, is it possible to optimize a given noise (rather than the model weights) to generate better results? Both answers are affirmative. Fortunately, we not only confirm the possibility but also propose practical algorithms. 

\begin{figure}[t]
\centering
\includegraphics[width =1.0\columnwidth ]{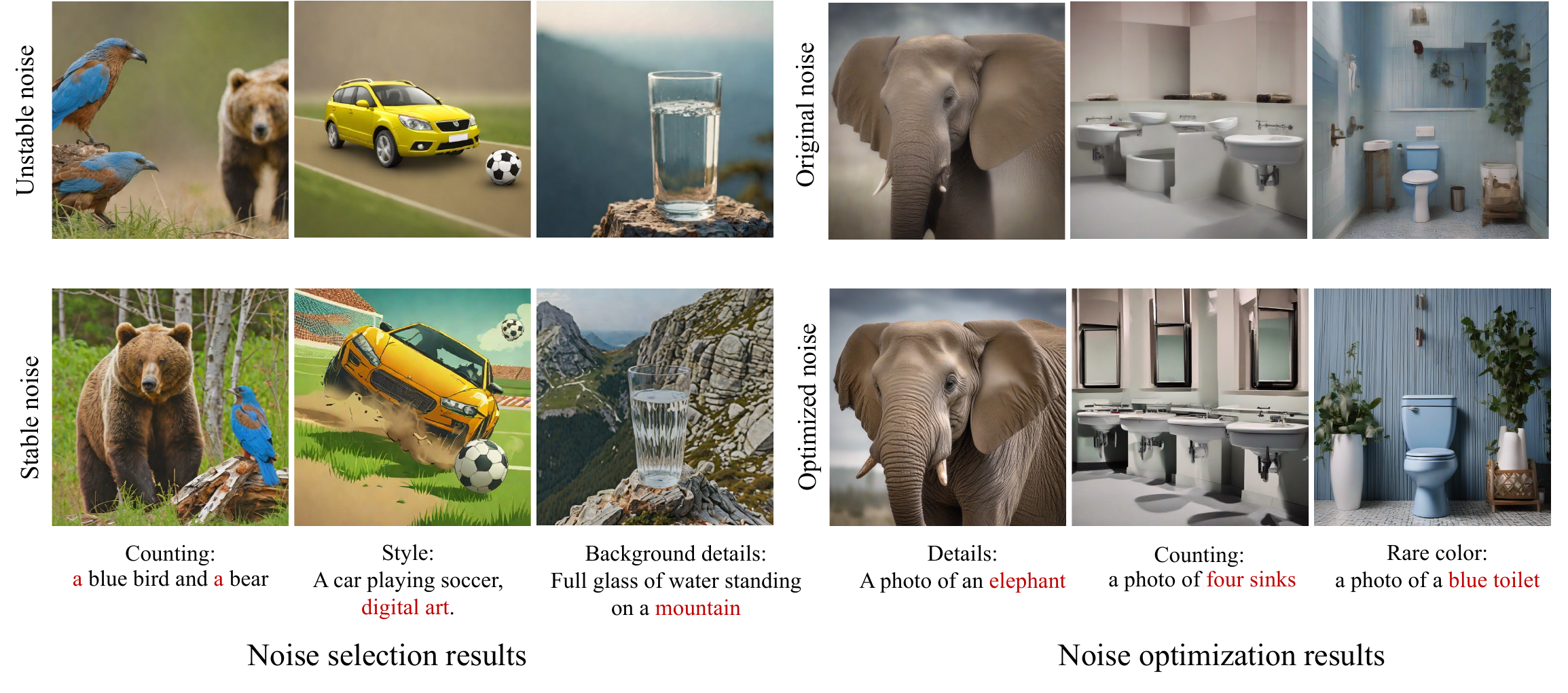}  
\caption{The qualitative results of noise selection and noise optimization. Left: SDXL-turbo. Right: SDXL. The proposed methods make improvements in multiple aspects.}
 \label{fig:slection_optimize_cases} 
\end{figure}

\textbf{Contributions.} We summarize the three main contributions of this work as follows: 

\textbf{First}, we are the first to hypothesize and empirically verify that not all noises are created equally. Specifically, random noises with high inversion stability usually lead to better generation than noises with lower inversion stability, where the inversion stability can be quantitatively given by the cosine similarity of the sampled noise $\epsilon$ and the inverse noise $\epsilon^{\prime}$. This quantitative metric naturally provides us with a novel noise selection method to select stable noises (e.g. the seed with the highest inversion stability among 100 noise seeds), which often correspond to better generated results. We present several qualitative results of noise selection in Figures \ref{fig:slection_optimize_cases} and \ref{fig:noise_selection}. 

\textbf{Second}, we further proposed a novel noise optimization method that actively enhances the inversion stability of arbitrary given noises. More specifically, we optimize an inversion-stability loss via gradient descent with respect to the sampled noise (rather than the convention model weight space). The proposed noise optimization method is the first one that works on noise space rather than model weight space to improve diffusion models. We present several qualitative results of noise optimization in Figures \ref{fig:slection_optimize_cases} and \ref{fig:random_optimized_pick&draw}. 

\textbf{Third}, our extensive experiments demonstrate that the proposed noise selection and noise optimization methods both significantly improve representative diffusion models, such as SDXL and SDXL-turbo. On one hand, the human preference winning rates of noise selection and noise optimization over the baseline can be up to $57\%$ and $72.5\%$, respectively, on DrawBench in terms of human preference. On the other hand, noise selection and noise optimization are also preferred by Human Preference Score (HPS) v2  \citep{wu2023human}, the latest state-of-the-art human preference model trained on diverse high-quality human preference data, with winning rates up to $67\%$ and $88\%$, respectively. Human preference, regarded as the ground-truth ultimate evaluation metric for text-to-image generation, and objective evaluation metrics generally support our methods.


\section{Prerequisites}
\label{sec:prerequisite}

In this section, we formally introduce prerequisites and notations.

\textbf{Notations.} Suppose a diffusion model $\mathcal{M}$ can generate a clean sample $x_0$ based on some condition $c$, such as a text prompt, given a sampled random noise $\epsilon$.\footnote{For simplicity, we abuse the latent space and the original data space in the presence of latent diffusion.} We denote the score neural network as $u_{\theta}(x_t,t)$, the model weights as $\theta$, the noisy sample at the $t$-th step as $x_t$, and $T$ as the total number of denoising steps.

\textbf{Diffusion Models.} The diffusion models  \citep{ho2020denoising} typically denoise a Gaussian noise along a reverse diffusion path (steps: $T \rightarrow 1$) to generate an image step by step. The probability via $u_{\theta}$, denoted as $p_\theta$, represents the sampling probability given the previous step's data. The starting point sampled from a Gaussian distribution, $p(x_T) = \mathcal{N}(x_T|\mathbf{0}, \mathbf{I})$. The probability of the whole chain, $p_\theta(x_{0:T})$, is shown as follows:
\begin{equation}
\small
\label{eq:denoising}
    p_\theta(x_{0:T}) = p(x_T)\prod^T_{t = 1} p_\theta(x_{t-1}|x_t), p_\theta(x_{t-1}|x_t) = \mathcal{N}(x_{t-1};\frac{1}{\sqrt{\alpha_t}}(x_t - \frac{\beta_t}{\sqrt{1-\overline{\alpha}_t}}u_\theta(x_t, t)), \sigma_{t}\mathbf{I}),
\end{equation}

where $a_t = 1 - \beta_t$, $\overline{\alpha}_t = \prod^s_{t=1}\alpha_s$. The $\beta_t$ and $\sigma_{t}$ are the pre-defined parameters for scheduling the scales of adding noises. The $\epsilon_{t}$ is an additional sampling noise at the $t$-th step.

\textbf{Noise Inversion.} The noise inversion is to invert a clean data into a noise along a pre-defined diffusion path. We can write the DDIM inversion process \citep{hertz2022prompt} as
\begin{equation}
\small
\label{eq:inversion}
     x_{t} \approx \sqrt{\frac{\alpha_{t}}{\alpha_{t-1}}} x_{t-1} + \sqrt{\alpha_{t-1} }(\sqrt{\frac{1-\alpha_{t}}{\alpha_t}} - \sqrt{\frac{1-\alpha_{t-1}}{\alpha_{t-1}}})u_\theta(x_{t-1}, t, c),
\end{equation}
where people approximate the denoising score prediction at $x_{t}$ with the inversion score prediction at $x_{t-1}$.
Equation \eqref{eq:denoising} can gradually transform a sampled noise $\epsilon$ into a generated sample $x_0$ along the denoising path, and Equation \eqref{eq:inversion} can gradually transform a generated sample $x_0$ back to a noise $\epsilon'$ along the noise inversion path. We note that the standard noising path which adds independent Gaussian noises is essentially different from the noise inversion path which adds the predicted noise of the score neural network $u_{\theta}$. While the generation denoising path and the noise inversion path are both guided by the score neural network $u_{\theta}$, the sampled noise $\epsilon$ and the inverse noise $\epsilon'$ are close but not identical due to the cumulative numerical differences.

\textbf{Fixed Points.} We denote the denoising-inversion transformation, $\epsilon \to x_0 \to \epsilon'$, as the transformation function $\epsilon' = F(\epsilon)$. If $\epsilon$ and $\epsilon'$ are ideally identical, namely $\epsilon = F(\epsilon)$, we call $\epsilon$ a fixed point of this mapping function $F$. In this case, the inverse noise $\epsilon'$ can perfectly recover the sample $x_0$ generated from $\epsilon$. This suggests that a state can remain fixed under some transformation. The fixed points have various great properties and many important applications in various fields, such as projective geometry \citep{coxeter1998non}, Nash Equilibrium \citep{nash1950equilibrium}, and Phase Transition \citep{wilson1971renormalization}.


\section{Methodology}
\label{sec:method}
In this section, we first introduce the noise inversion stability hypothesis and show how it naturally leads to two novel noise-space algorithms, including noise selection and noise optimization.
\begin{figure}[t]
\centering
\includegraphics[width =1.0\columnwidth ]{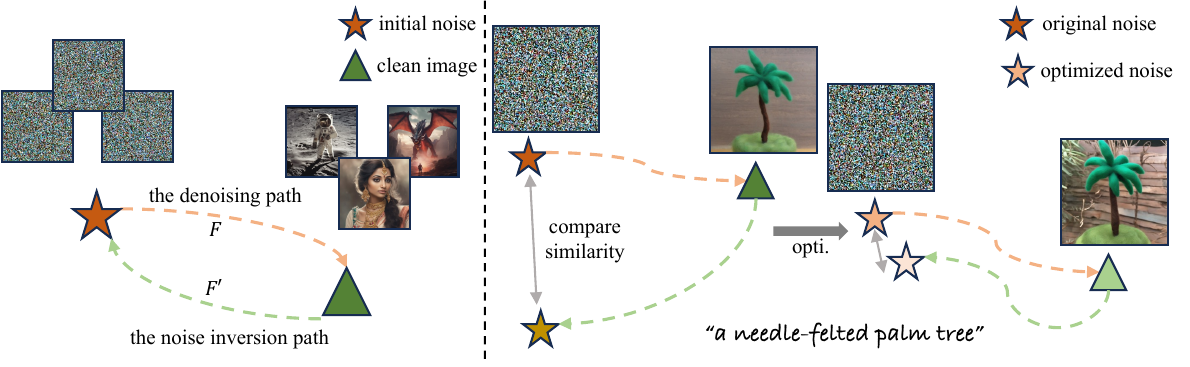}  
\caption{The overview of noise inversion stability and noise optimization. Left: an ideal fixed-point noise is the state that remains unchanged through a cyclic transformation of denoising and inversion. Right: random noises are not perfect fixed-point noises for the denoising-and-inversion transformation, which lead to some difference between original noises and inverse noises. We can select stable noises or directly optimize the given random noises to get closer to a fixed point.}
 \label{fig:overview} 
\end{figure}

\begin{minipage}{0.45\textwidth}
    \begin{algorithm}[H]
    \small
    \caption{Noise Selection}
    \begin{algorithmic}[1]
    \STATE {\bfseries Input:} the diffusion model: $\mathcal{M}$, text prompt: $c$, the number of seeds: $K$
    \STATE {\bfseries Output:} the stable noise $\epsilon_s$
    \FOR{$i=1$ {\bfseries to} $K$}
       \STATE $\mathrm{seed} \leftarrow i$ // Set the random seed
       \STATE Sampling a Gaussian noise $\epsilon_{i}$
       \STATE $x_0 = \mathcal{M}(\epsilon_{i}, c)$ \\
       // Generate an image, Equation~\eqref{eq:denoising}
       \STATE $\epsilon'_{i} = \mathrm{Inversion}(x_0, c)$ \\
       // Inverse noise, Equation~\eqref{eq:inversion}
       \STATE $\mathrm{s}(\epsilon_i) = \cos(\epsilon_{i}, \epsilon'_{i})$ 
    \ENDFOR
    \STATE  $\epsilon_s =\argmax\limits_{\epsilon \in \{\epsilon_i | i = 1, 2, \cdots, K\}}\ s(\epsilon)$ \\
    // The noise with the highest stability score
    \end{algorithmic}
    \label{algo:noise_selection}
    \end{algorithm}
\end{minipage}
\hfill
\begin{minipage}{0.45\textwidth}
    \begin{algorithm}[H]
    \small
    \caption{Noise Optimization}
    \begin{algorithmic}[1]
    \STATE {\bfseries Input:} the diffusion model: $\mathcal{M}$, text prompt: $c$, the number of gradient descent steps: $n$, the learning rate: $\eta$, the momentum value: $\beta$
    \STATE {\bfseries Output:} the optimized noise $\epsilon^{\star}$
    \STATE Sampling a Gaussian noise $\epsilon$
   \FOR{$i=1$ {\bfseries to} $n$}
       \STATE $x_0 = \mathcal{M}(\epsilon, c)$ \\
        // Generate an image, Equation~\eqref{eq:denoising}
       \STATE $\epsilon' = \mathrm{Inversion}(x_0, c)$ \\
       // Inverse noise, Equation~\eqref{eq:inversion}
       \STATE $J(\epsilon) = 1 - \cos(\epsilon, \epsilon')$
       \STATE $m_i = \beta m_{i-1} + \nabla_{\epsilon} J(\epsilon)$ 
       \STATE $\epsilon = \epsilon - \eta  m_i$
   \ENDFOR
   \\
   $\epsilon^{\star} = \epsilon$
    \end{algorithmic}
    \label{algo:noise_optimization}
    \end{algorithm}
\end{minipage}

\textbf{Noise Inversion Stability.} It is well known that fixed points are stable under the transformation and, thus, have great properties~\citep{burton2003stability, connell1959properties, pata2019fixed}. May the fixed-point Gaussian noises under the denoising-inversion transformation $F$ also exhibit some advantages? As finding the fixed points of this complex dynamical system is intractable, unfortunately, we cannot empirically verify it. Instead, we can formulate Definition \ref{df:noise_stability} to measure the stability of noise for the denoising-inversion transformation $F$.
\begin{define}[Noise Inversion Stability]
 \label{df:noise_stability}
Suppose a sampled noise $\epsilon$ has its inverse noise $\epsilon' = F(\epsilon)$ for the denoising-inversion transformation $F$ given by a diffusion model $\mathcal{M}$ with the condition $c$. We define the noise inversion stability of the sampled noise $\epsilon$ as
\begin{equation}
\label{eq:stability}
 s(\epsilon) = \cos(\epsilon, \epsilon')
\end{equation}
for the diffusion $\mathcal{M}$ with the condition $c$, where $\cos$ is the cosine similarity between two vectors.
\end{define}
We use cosine similarity to measure stability for simplicity, while it is also possible to use other similarity metrics. Our empirical analysis in Section \ref{sec:empirical} suggests that the simple cosine similarity metric works well.

\textbf{Noise Selection.} Inspired by the intriguing mathematical properties of fixed points, we hypothesize that the noise with higher inversion stability can lead to better generated results. If this hypothesis is reasonable, this naturally provides a novel and useful noise selection algorithm that selects the noise seed with the highest stability score from $K$ noise seeds (e.g. $K=100$ in this work). We present the pseudocode in Algorithm \ref{algo:noise_selection}.

\textbf{Noise Optimization.} As we have an objective to increase the noise inversion stability, is it possible to actively optimize a given noise by maximizing the stability score? 
We further propose the noise optimization algorithm that directly performs Gradient Descent (GD) on the loss, $1-\cos(\epsilon, \epsilon')$, with respect to $\epsilon$, where we keep the diffusion model weights and $\epsilon'$ constant for each optimization step. We present the illustration of noise optimization in the right column of Figure \ref{fig:overview}. We present the pseudocode in Algorithm \ref{algo:noise_optimization}.

\section{Empirical Analysis}
\label{sec:empirical}
In this section, we conduct extensive experiments to demonstrate the effectiveness of our methods. We take text-to-image generation as our main setting. 

\subsection{Experimental Settings}
\textbf{Models:} SDXL-turbo  \citep{sauer2023adversarial} and SDXL \citep{podell2023sdxl}. SDXL is a representative and powerful diffusion model. SDXL-turbo is a recent accelerated diffusion model which can produce results better than standard SDXL but only take 4 denoising steps. We choose the denoising steps for SDXL-turbo as 4 steps and SDXL as 10 steps for reducing computational time and carbon emissions, unless we specify it otherwise. We also empirically study how the proposed methods depend on the denoising steps in Appendix \ref{sec:suppexp}. 

\textbf{Dataset:}  We use all 200 test prompts from the DrawBench dataset \citep{saharia2022photorealistic} which contain comprehensive and diverse descriptions beyond the scope of the common training data. We use the first 100 test prompts from the Pick-a-Pic \citep{kirstain2024pick} which consist of interesting prompts gathered from the users of the Pick-a-Pic web application. In case studies on color, style, text rendering, object co-occurrence, position, and counting, we specifically collected some prompts from the GenEval dataset \citep{ghosh2024geneval}.


\textbf{Evaluation metrics:} We evaluate the quality of the generated images using both human preference and popular objective evaluation metrics, including HPS v2 \citep{wu2023human}, AES \citep{schuhmann2022laion}, PickScore \citep{kirstain2024pick}, and ImageReward \citep{xu2024ImageReward}. AES indicates a conventional aesthetic score for images, while HPS v2, PickScore, and ImageReward are all emerging human reward models that approximate human preference for text-to-image generation. Particularly, HPS v2 is the state-of-the-art human reward model so far and offers a metric more close to human preference (see Table 6 in ~\citep{wu2023human}) than other objective evaluation metrics. Moreover, human preference is regarded as the ground-truth and ultimate evaluation method for text-to-image generation. Thus, we regard human preference and HPS v2 as the two most important metrics.

\textbf{Hyperparameters:} For the noise selection experiments, we select the (most) stable noise and the (most) unstable noise from 100 noise seeds according the noise inversion stability. We evaluate generated results using human preference and objective evaluation metrics. For the noise optimization experiments, we initialize the noise using one random seed and perform GD to optimize the noise with 100 steps. The default values of the learning rate and the momentum are 100 and 0.5, respectively. 


\subsection{The Experiments of Noise Selection}
\label{sec:noise_selection}

\begin{table}[t]
\caption{The quantitative results of noise selection. Each reported score is the mean score over all evaluated prompts. The corresponding winning rate results are shown in Figure \ref{fig:noise_selection_rate} and the qualitative results are shown in Figure \ref{fig:noise_selection}. Model: SDXL-turbo.}
\label{table:noise_selection}
\centering
\begin{tabular}{c|c|cccc|c}
\toprule
Dataset  & Noise  & HPS v2 & AES & PickScore  & ImageReward  & Average\\
\midrule
\multirow{2}{*}{Pick-a-Pic} & Unstable noise & 27.2688  & 5.9265 &  21.6227 & 0.7812  &13.8998\\
  & Stable noise  & \textbf{27.4934}  & \textbf{5.9960} & \textbf{21.6372}  &  \textbf{0.8981} & \textbf{14.0062}  \\
\midrule
\multirow{2}{*}{DrawBench} & Unstable noise   & 28.1377 & 5.3945 & \textbf{22.4251} & 0.7021 & 14.1646\\
& Stable noise  & \textbf{28.4266}  & \textbf{5.6082}  & 22.4200 & \textbf{0.7325} & \textbf{14.2968}\\
\bottomrule
\end{tabular}
\end{table}

\begin{figure}[t]
\centering
\includegraphics[width =1.0\columnwidth ]{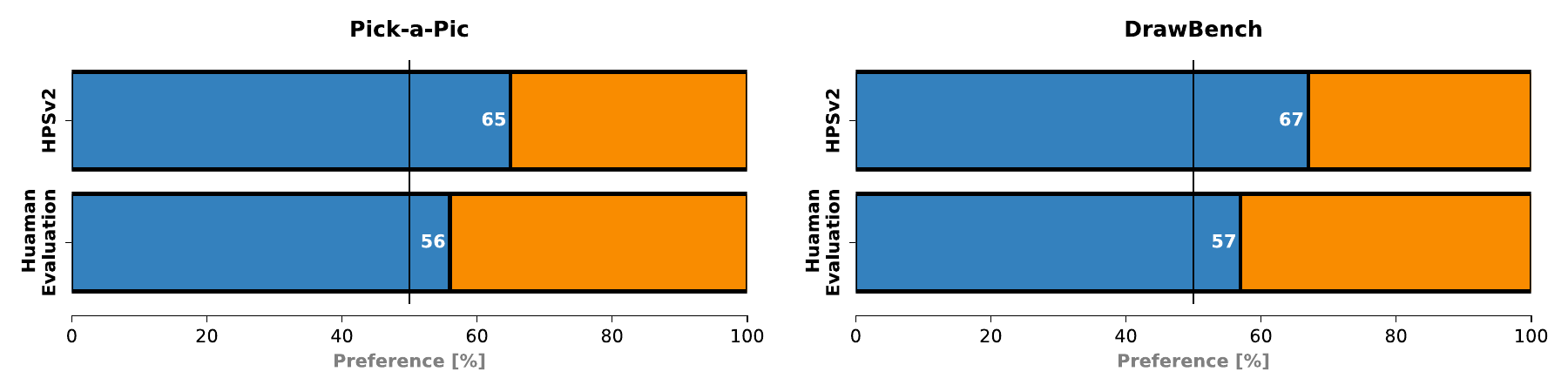}  
\caption{The winning rate results of noise selection. The blue bars represent the side of stable noises. The orange bars represent the side of unstable noises. Mode: SDXL-turbo.}
 \label{fig:noise_selection_rate} 
\end{figure}

\begin{figure}[t]
\centering
\includegraphics[width =1.0\columnwidth ]{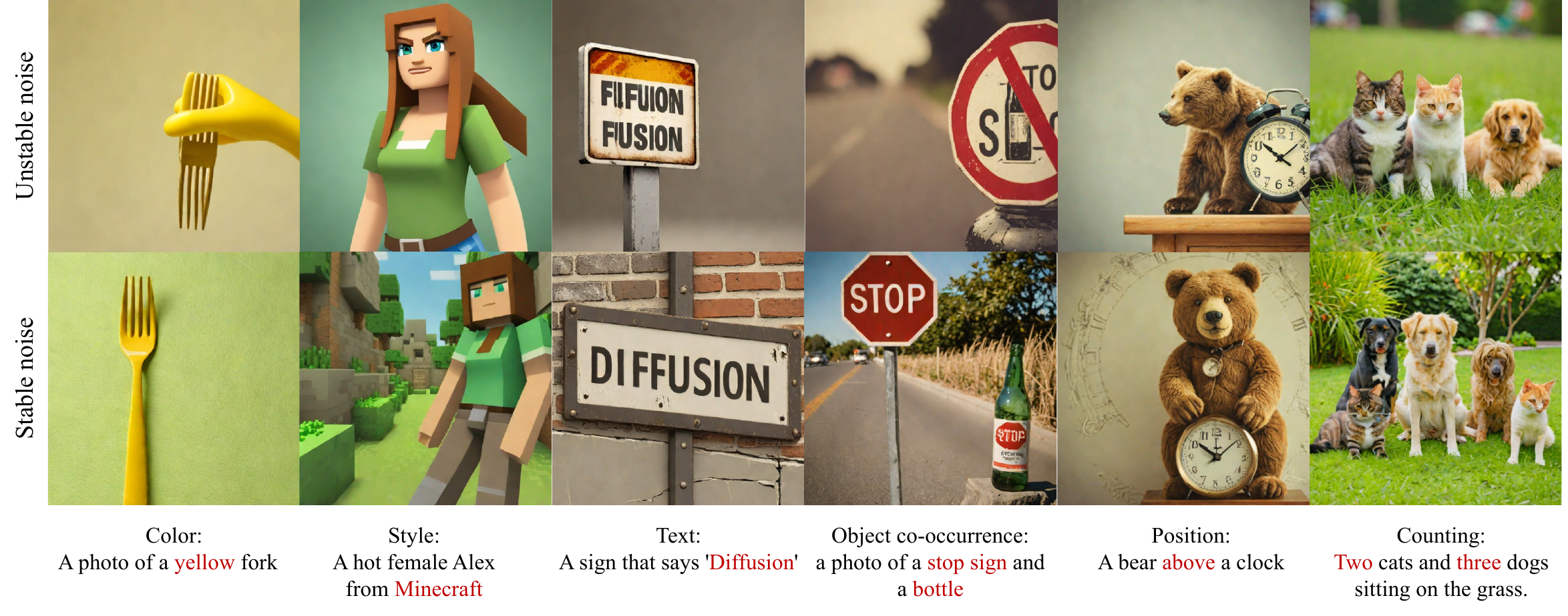}  
\caption{The qualitative results of noise selection. The results highlight the improvements of stable noises in various aspects, such as color, style, text rendering, object co-occurrence, position, and counting. The prompts are from Pick-a-Pic, DrawBench and GenEval datasets. Model: SDXL-turbo.}
 \label{fig:noise_selection} 
\end{figure}

The noise selection experiments are to compare the results denoised from stable noises and unstable noises, where the noise with the highest stability score is the stable noise and the noise with the lowest stability score score is defined as unstable noise. 

\textbf{Quantitative results.} We present the objective evaluation scores in Table \ref{table:noise_selection}.
The HPS v2 is the main objective evaluation metric, as it is the state-of-the-art human reward model. The HPS v2 score of stable noises surpasses it counterpart of unstable noise by 0.225 and 0.289, respectively, on Pick-a-Pic and DrawBench. The average scores also support the advantage of stable noises over unstable noises. The quantitative results supports the noise inversion stability hypothesis and suggest that stable noises often significantly outperform unstable noises in practice.

Besides the scores, the winning rates can tell the percentage of one result better than the other on the evaluated prompts. We particularly show the winning rates of human preference and and HPS v2 in Figure \ref{fig:noise_selection_rate} to visualize two representative evaluation metrics. All winning rates are significantly higher than 50\%. The human preference winning rates are higher than \textbf{56\%}. The HPS v2 winning rates are even up to \textbf{65\% and 67\%} over Pick-a-Pic and DrawBench.

\textbf{Qualitative results.} 
We conduct case studies for qualitative comparison. We not only care about the standard visual quality, but also further focus on those challenging cases for diffusion models, such as color, style, text rendering, object co-occurrence, position, and counting. The results in Figure \ref{fig:noise_selection} show that the images denoised from stable noise are significantly better than images denoised from unstable noise in various aspects. 
1) Color: the stable noise leads to a yellow fork accurately, while the unstable noise can only lead to a yellow hand with an incorrect fork. 2) Style: the stable noise obviously correspond to the ``Minecraft'' style more precisely with rich background details. 3) Text rendering: the stable noise can render the correct ``diffusion''. 4) Object co-occurrence, the stable noise can generate correct combinations of two objects, while the unstable noise falsely merges two concepts together. 5) Position, the stable noise correct the wrong position relation of the unstable noises. 6) Counting, the stable noises accurately correct the number of both cats and dogs.

In summary, both quantitative and qualitative results demonstrate the significant effectiveness of noise selection according to the noise inversion stability.

\subsection{The Experiments of Noise Optimization} 
\label{sec:noise_optimization}

\begin{table}[t]
\caption{The quantitative results of noise optimization. The qualitative results are shown in Figure \ref{fig:random_optimized_pick&draw}, and the winning rate results are shown in Figure \ref{fig:noise_optimize_rate}. Model: SDXL.}
\label{table:noise_optimize_pick&draw}
\centering
\begin{tabular}{c|c|cccc|c}
\toprule
Dataset & Noise & HPS v2 & AES & PickScore & ImageReward  & Average\\
\midrule
\multirow{2}{*}{Pick-a-Pic} & Original Noise  & 25.9800 & 5.9903   & 21.0183 & 0.2500 & 13.3207\\
& Optimized Noise   & \textbf{26.6422}  & \textbf{6.0504} & \textbf{21.2344} & \textbf{0.4622} & \textbf{13.5973}\\
\midrule
\multirow{2}{*}{DrawBech} & Original Noise  & 26.6203 & 5.4889 & 21.4815  & 0.0575 & 13.4121\\
& Optimized Noise  & \textbf{27.3651}  & \textbf{5.5438} & \textbf{21.6508}  & \textbf{0.1767} & \textbf{13.6841}\\ 
\bottomrule
\end{tabular}
\end{table}

\begin{figure}[t]
\centering
\includegraphics[width =1.0\columnwidth ]{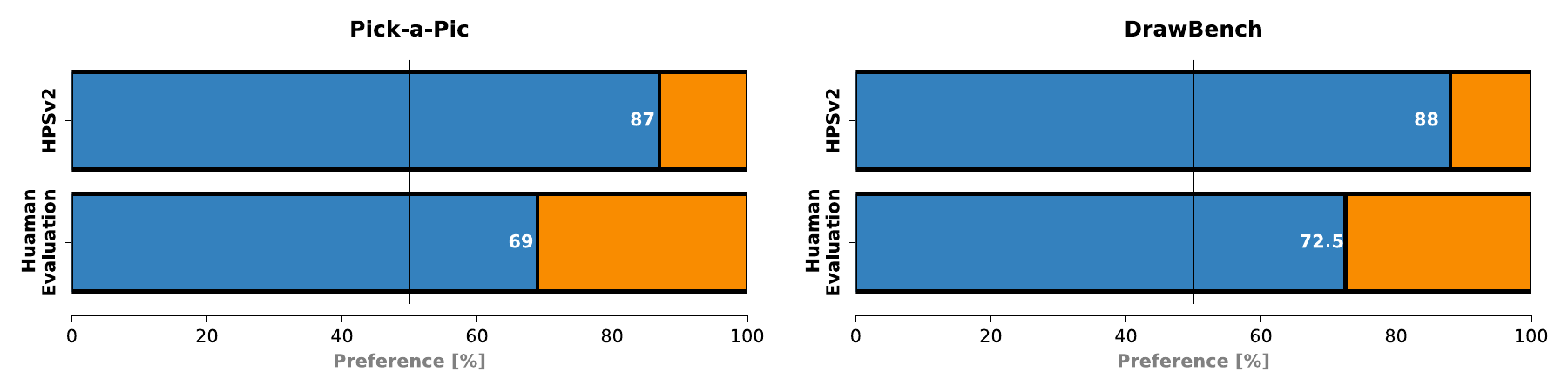}  
\caption{The winning rate result of noise optimization. The blue bars represent the side of optimized noises. The orange bars represent the side of the original noise. Model: SDXL.}
 \label{fig:noise_optimize_rate} 
\end{figure}

\begin{figure}[t]
\centering
\includegraphics[width =1.0\columnwidth ]{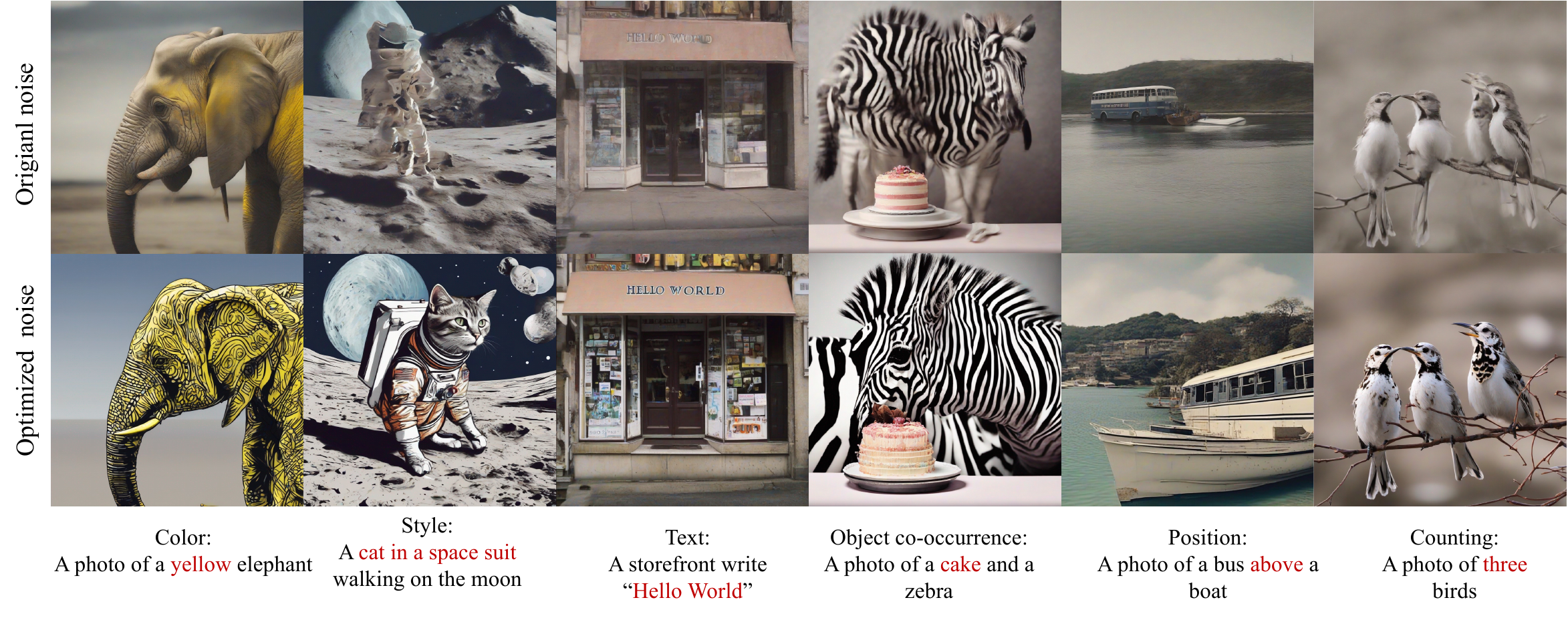}  
\caption{ The qualitative results of noise optimization on Pick-a-Pic, DrawBench and GenEval datasets. Each pair of results is generated by SDXL. The results demonstrate that the optimized noise outperforms the original noise in various aspects, such as color, style, text rendering, object co-occurrence, position, and counting.}
 \label{fig:random_optimized_pick&draw} 
\end{figure}

\begin{figure}[t]
\centering
\includegraphics[width =1.0\columnwidth ]{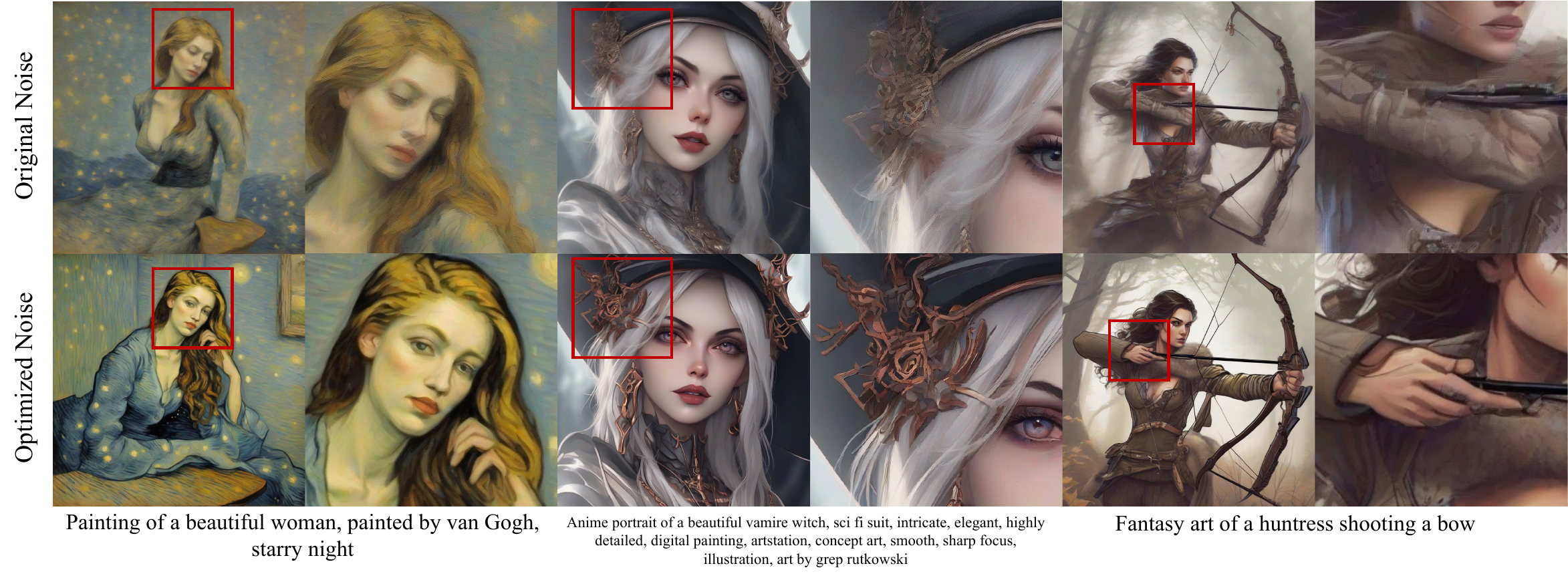}  
\caption{The character and body details of original noises and optimized noises. The prompts are form Pcik-a-Pic. Model: SDXL}
 \label{fig:human_zoom_in} 
\end{figure}

\begin{figure}[t]
\centering
\includegraphics[width =1.0\columnwidth ]{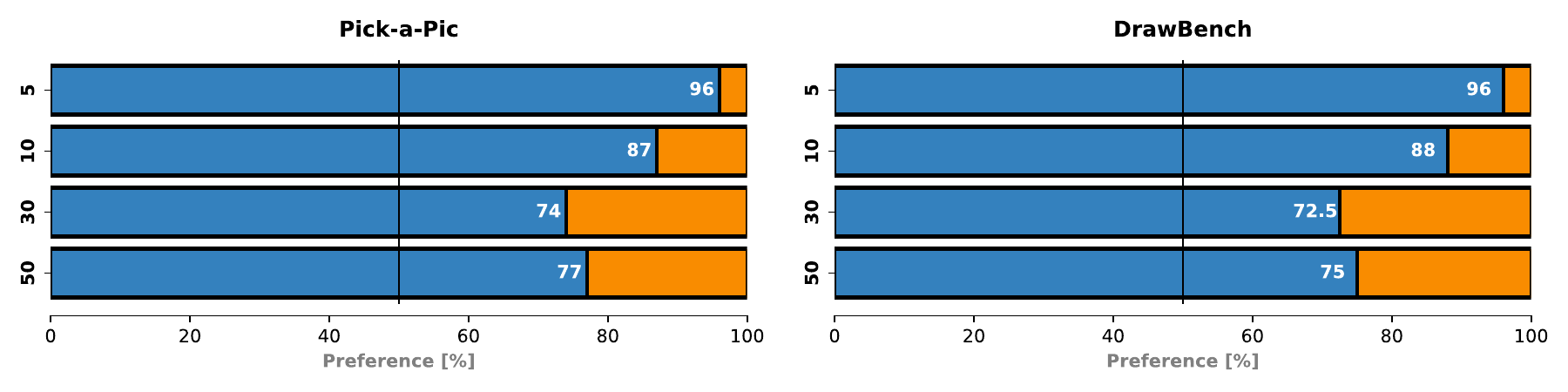}  
\caption{The winning rates of noise optimization with respect to various denoising steps. Metric: HPS v2, Model: SDXL.}
 \label{fig:HPS-rate-5-50-pick&draw} 
\end{figure}

The noise optimization experiments are to compare the results of original noises and optimized noises. For each prompt, we sample a Gaussian noise as the original noise and learn optimized noises by Algorithm \ref{algo:noise_optimization}. Note that optimized noises are approximately but not real Gaussian noises.  

\textbf{Quantitative results.} We present quantitative results in Table \ref{table:noise_optimize_pick&draw}. All objective evaluation metrics in the experiment consistently support the advantage of optimized noises over original noises. The HPS v2 score of optimized noises surpasses it counterpart of original noises by 0.662 and 0.745, respectively, on Pick-a-Pic and DrawBench. The average score again supports the advantage of optimized noises over original noises.

Similarly, we visualize the winning rates of two most important metrics, HPSv2 and human preference to show the percentage of improved cases in Figure \ref{fig:noise_selection_rate}. The human preference winning rates of noise optimization are \textbf{69\% and 72.5\%}, respectively, over Pick-a-Pic and DrawBench, while the HPS v2 winning rates are even up to \textbf{87\% and 88\%}. The winning rate improvements are comparable to the performance gap between two generations of SD models, such as SDXL-turbo~\citep{sauer2023adversarial} and cascaded
pixel diffusion models (IF-XL)~\citep{saharia2022photorealistic} . 


\textbf{Qualitative results.} We presents the qualitative results of original noises and optimized noises in Figure \ref{fig:random_optimized_pick&draw}. Similar to what we observe for noise selection, noise optimization also improve multiple challenging cases, such as color, style, text rendering, object co-occurrence, position, and counting we mentioned above. Moreover, we also present the examples that noise optimization can improve the details of human characters and bodies in Figure~\ref{fig:human_zoom_in}. Optimized noises can lead to more accurate human motion and appearance. For example, the huntress's hand generated by the optimized noise are accurately holding the end of the arrow.


\textbf{Robustness to the number of denoising steps $T$.} The noise inversion process directly depend on the number of denoising steps $T$. We apply our noise optimization to SDXL with various denoising steps to study the robustness of noise optimization to the hyperparameter $T$. We present the winning rates of noise selection with $T \in \{5, 10, 30, 50\}$ in Figure \ref{fig:HPS-rate-5-50-pick&draw}. The results shows that the improvement of noise optimization is relatively robust to a wide choice of denoising steps. Optimized noises are especially good for very few denoising steps.

\begin{figure}[t]
\centering
\includegraphics[width =1.0\columnwidth ]{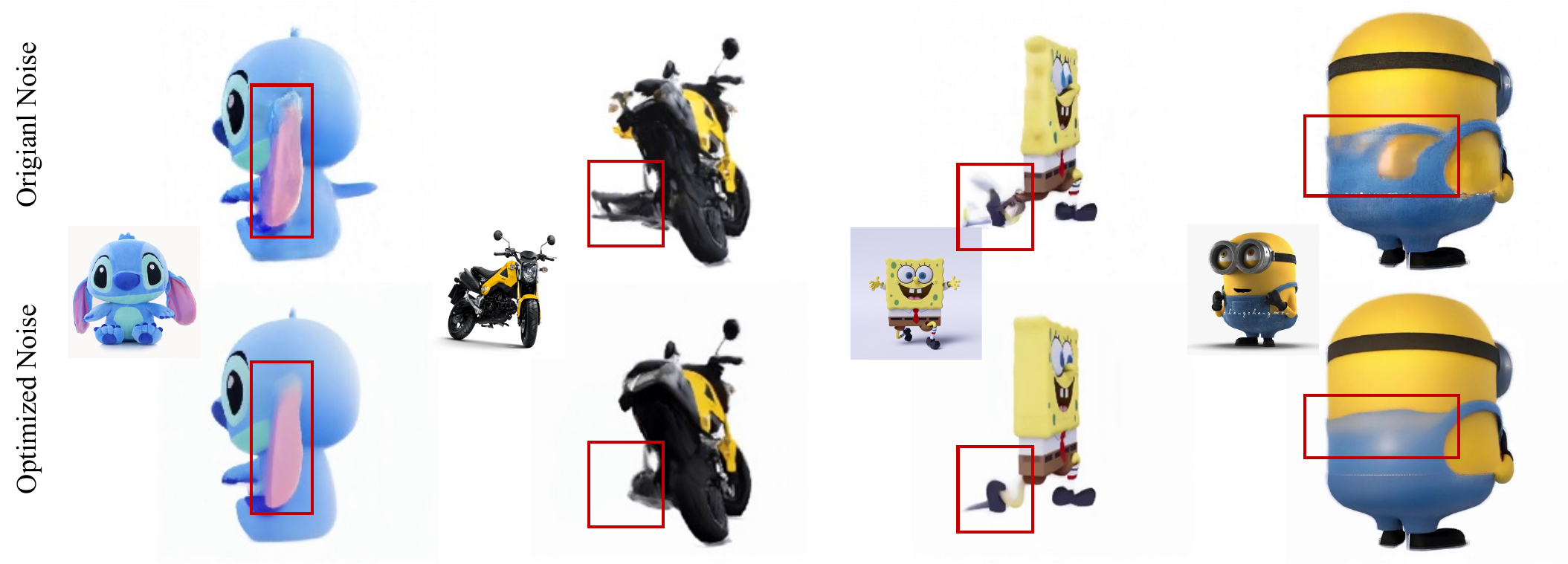} 
\vspace{-0.3mm}
\caption{The qualitative results of noise optimization for 3D generation. The small images are the input. The red box highlights the differences. Model: SV3D.}
 \label{fig:case_study_3d_0} 
\end{figure}

\textbf{Noise Optimization for 3D Generation.} It is easy to see that the proposed methods can be generally applied to other diffusion models. Here, we provide an example. We apply noise optimization to 3D generation tasks with a popular image-to-3D generative model, SV3D \citep{voleti2024sv3d}. We clearly observe the improvements in body details of these 3D characters. Due to the space limit, we leave more experimental details and results in Appendix \ref{sec:3d}.

In summary, noise optimization can significantly improve generated results in multiple challenging aspects. It is especially surprising that optimized noises deviated from Gaussian noises can help diffusion models generate better results than real Gaussian noises.





\section{Discussion and Limitations}
\label{sec:discuss}

In this section, we further discuss related works and the limitations of this work. While this work reports very interesting findings and proposes novel algorithms on noise space of diffusion models, it still has several limitations. 


\textbf{Related Work} The noise inversion technique is mainly applied on image editing \citep{mokady2023null, meiri2023fixed, huberman2023edit} in very similar ways. They usually invert a clean image into a relatively noisy one via a few inverse steps and then denoise the inverse noisy images with another prompt to achieve instruction editing. Some works \citep{mao2023guided} in this line of research realized that editing noises can help editing generated results. Specifically, modifying a portion of the initial noise can affect the layout of the generated images. Other works \citep{liu2024drag, shi2023dragdiffusion} focused on dragging and dropping image content via interactive noise editing. However, the goal of previous studies is to control image layout under fine-grained control conditions, such as input layout or editing operations. In contrast, we focus on generally improving generated results of diffusion models by selecting or optimized a Gaussian noise according to the stability score. 

\textbf{Theoretical Understanding.} With the inspirations from fixed points in dynamical systems, we still do not theoretically understand why not all noises are created for diffusion models. We formulated and empirically verified the hypothesis that random noises with higher inversion stability often lead to better generated results, it is still difficult to theoretically analyze how the performance of diffusion models mathematically depend on noise stability. We believe theoretically understanding noise selection and noise optimization will be a key step to further improve them. 

\textbf{Optimization Strategies.} In this, we only applied simple gradient descent with multiple (e.g., $100$) steps to optimize the noise-space loss, but noise optimization seems like a difficult optimization task. In some cases, we observe that the loss does not converge smoothly. Due to computational costs and poor understanding towards noise-space loss landscape, we did not carefully fine-tune the hyperparameters or employ advanced optimizers, such as Adam \citep{kingma2014adam} in this work. Thus, our current optimization strategy is far from releasing the power of noise-space algorithms. We think it will be very promising and important to better analyze and solve this emerging optimization task with advanced methods.

\textbf{Computational Costs.} Both noise selection and noise optimization require significantly more computational costs and time than the standard generation. For noise selection, we compute the inversion stability loss of 100 noise seeds and select the one with the highest stability score. Thus, we need to repeat the forward and inversion process for 100 times. For noise optimization, we perform gradient descent with 100 steps. Thus, we need to repeat the forward, inversion and gradient computing process each for 100 times. It may be difficult to specifically accelerate each step of noise selection, but it will be very likely to accelerate noise optimization with less GD steps in near future. 

\section{Conclusion}
\label{sec:conclusion}
In this paper, we report an interesting noise inversion stability hypothesis and empirically observe that noises with higher inversion stability often lead to better generated results. This hypothesis motivates us to design two novel noise-space algorithms, noise selection and noise optimization, for diffusion models. To the best of our knowledge, we are the first to report that not all noises are created equally for diffusion models and the first to generally improve diffusion models without fine-tune the model parameters. Our extensive experiments demonstrate that the proposed methods can significantly improve multiple aspects of qualitative results and enhance human preference rates as well as objective evaluation scores. Moreover, the proposed methods can be generally applied to various diffusion models in a plug-and-play manner. While some limitations exist, our work has made the first solid step to explore this promising direction. We believe our work will motivate more studies on understanding and improving diffusion models from the perspective of noise space.

\bibliography{deeplearning}

\newpage
\appendix
 
\section{Experimental Settings of Main Experiments}
\label{sec:expsetting}

\textbf{Computational environment.} The experiments are conducted on a computing cluster with GPUs of NVIDIA\textsuperscript{\textregistered} Tesla\textsuperscript{\texttrademark} A100.

\subsection{Datasets and Data Preprocessing}
\label{sec:setting}
we conduct experiments across three datasets as follows:

\textbf{Pick-a-Pic.}~\citep{kirstain2024pick}: This dataset is composed of data collected from users of the Pick-a-Pic web application. Each example in this dataset consists of a text prompt, a pair of images, and a label indicating the preferred image. It is worth noting that for fast validation and saving computational resources, we only use the first 100 prompts as text conditions to generate images in the main experiment.

\textbf{DrawBench.}~\citep{saharia2022photorealistic}: The examples in this dataset contain a prompt, a pair of images, and two labels for visual quality and prompt alignment. The total number of examples in this dataset is approximately 200. This dataset contains 11 categories of prompts that can be used to test various properties of generated images, such as color, number of objects, text in the scene, etc. The prompts also contain long, complex descriptions, rare words, etc. 

\textbf{GenEval.}~\citep{ghosh2024geneval}: This dataset contains 553 prompts spanning six attributes, including single object, two objects, counting, colors, positions, and attribute binding. The text prompts are generated from the templates\footnote{Templates and prompts are from https://github.com/djghosh13/geneval/tree/main/prompts}.

\textbf{The difference between these datasets}: The prompts in Pick-a-Pic are from real users and have more daily descriptions. The prompts in DrawBench have more complex descriptions and contain rare words. The prompts in GenEval are simple but outstanding property descriptions.

In all main experiments, we set all tensor as half precision to improve experimental efficiency. In calculating the inversion stability, we expand the noise tensor to a one-dimensional vector along the channel dimension.

\subsection{The Hyperparameters:}

\textbf{Noise Selection.} In noise selection experiments, for each prompt, we sample 100 noises using random seeds from 0 to 99. According to the inversion stability score, we select the stable noise among all candidate noises, using the algorithm~\ref{algo:noise_selection}.

\textbf{Noise Optimization.} In noise optimization experiments, for each prompt, we first randomly sample a noise using a random seed selected from 0 to 99. This noise is denoted as original noise. We use algorithm~\ref{algo:noise_optimization} to optimize the original noise with 100 gradient descent steps. We set the defaulted learning rate is 100 and equip it the learning rate with a cosine annealing schedule. The default value of monument is 0.5.

\subsection{Evaluation Metrics}

\textbf{Human Preference Score v2 (HPS v2)}: This score is calculated by a finetuned CLIP\footnote{The CLIP version is ViT-H/14} on the HPD v2 dataset~\citep{wu2023human}, a comprehensive human preference dataset. This human preference dataset is known for its diversity and representativeness. Each instance in the dataset contains a pair of images with prompt and a label of human preference. 

\textbf{Aesthetic Score (AES)}: The AES\footnote{The Github page: https://github.com/christophschuhmann/improved-aesthetic-predictor} is calculated by the Aesthetic Score Predictor~\citep{schuhmann2022laion}, which is  designed by adding five MLP layers on top of a frozen CLIP\footnote{The CLIP version is ViT-H/14} and only the MLP layers are fine-tuned by a regression loss term on SAC~\citep{pressmancrowson2022}, LAION-Logos\footnote{https://laion.ai/blog/laion-aesthetics/} and AVA~\citep{murray2012ava} datasets. The score is ranged from 0 to 10. A higher score means the image has better visual quality.

\textbf{PickScore}: This is also a human preference model. The score is calculated by a finetuned CLIP which is trained on the Pick-a-Pic dataset with a large number of user-annotated human preference data samples. 

\textbf{ImageReward}: This is an early human preference model~\citep{xu2024ImageReward}.

\begin{figure}[h]
\centering
\includegraphics[width =0.8\columnwidth ]{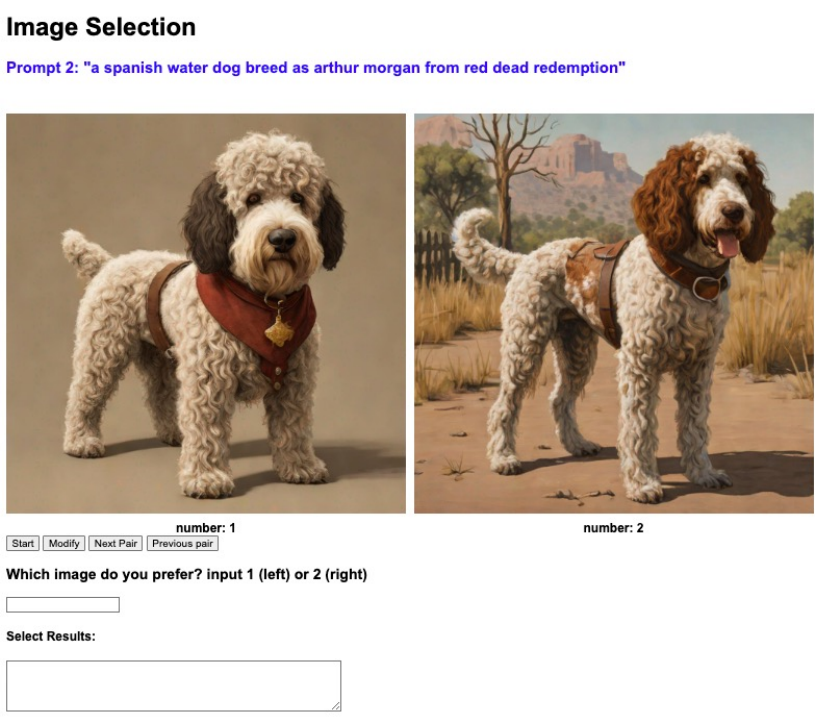}  
\caption{The web page for human evaluation.}
 \label{fig:usr_study} 
\end{figure}

\textbf{Human evaluation}: Human annotators select a better one from a pair of images following the criteria:
\begin{itemize}
    \item The correctness of semantic alignment
    \item The correctness of object appearance and structure
    \item The richness of details
    \item The aesthetic appeal of the image
    \item Your preference for upvoting or sharing it on social networks
\end{itemize}
We built a web page for human evaluation, as shown in Figure \ref{fig:usr_study}. 

\textbf{The difference between these metrics}: The AES is primary for evaluating the visual quality, while others are for human preference.

\newpage
\section{Supplementary Experimental Results}
\label{sec:suppexp}



\begin{figure}[h]
\centering
\includegraphics[width =1.0\columnwidth ]{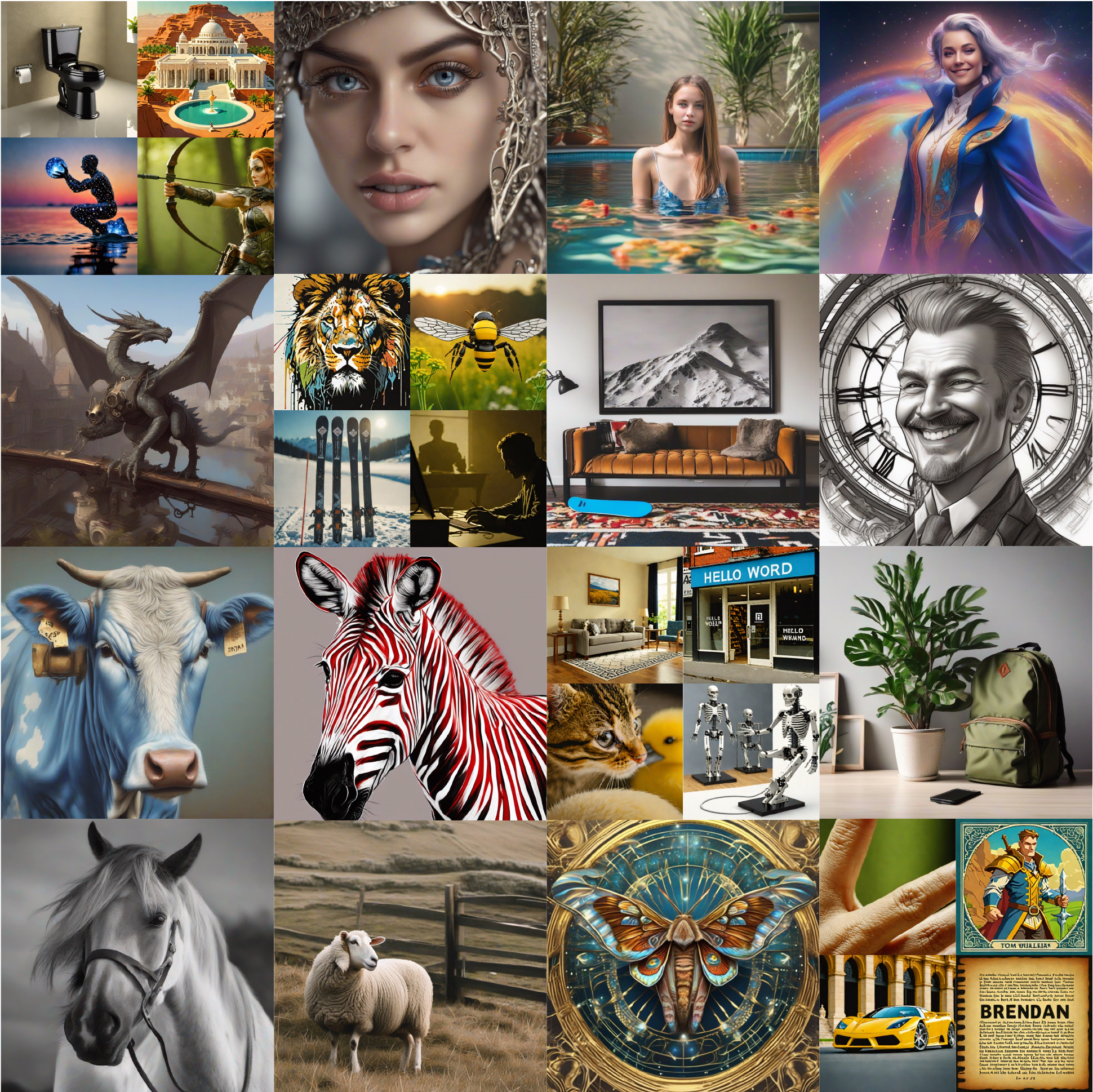}  
\caption{More results of optimized noises. The large images are generated by SDXL, and small images are generated by SDXL-turbo.}
 \label{fig:optimized_cases} 
\end{figure}

We show more results of noise optimization experiments in Figure~\ref{fig:optimized_cases}

\newpage
\section{3D Object Generation}
\label{sec:3d}

In this section, we analyze noise optimization for 3D diffusion models. 

\subsection{Methodology} 

The noise inversion rule of image-to-3D diffusion models is different from text-to-image diffusion. Here we derive the noise inversion rule for the popular image-to-3D diffusion model, SV3D \citep{voleti2024sv3d}.

SV3D employs the EDM framework \citep{karras2022elucidating}, which improves upon DDIM with a reparameterized to the denoising process. Taking a single image as input, SV3D generates a multi-view consistent video sequence of the object based on a specified camera trajectory, showcasing remarkable spatio-temporal properties and generalization capabilities. Specifically, we choose the SV3D-U variant, which, during training, consistently conditions on a static trajectory to generate a 21-frame 3D video sequence, with each frame representing a 360/21 degree rotation of the object.

The denoising process $x_{t+1} \rightarrow x_{t}$ within the EDM framework can be written as
\begin{equation}
\label{eq: edm_denoise_0}
\mathrm{x_t} = \mathrm{x_{t+1}} + \frac{\sigma_{t} - \sigma_{t+1}}{\sigma_{t+1}}\mu,
\end{equation}
\begin{equation}
\label{eq: edm_denoise_1}
\mu = x_{t+1} - \left(c_{skip}^{t+1}x_{t+1} + c_{cout}^{t+1}u_{\theta}(c_{in}^{t+1}\hat{x}_{t+1}; c_{noise}^{t+1})\right).
\end{equation}
We denote $\sigma_{t}$ as the noise level of the scheduler at the $t$-th time step and $u_{\theta}$ denotes the score network. $c_{skip}$, $c_{out}$, $c_{in}$, and $c_{noise}$ are coefficients dependent on the noise schedule and the current time step $t$ in Euler sampling method. Subsequently, if we intend to achieve noise inversion $\hat{x}_t$ $\rightarrow$ $\hat{x}_{t+1}$, we can modify Equation $\eqref{eq: edm_denoise_0}$ accordingly as
\begin{equation}
\label{eq: edm_inversion_0}
\hat{x}_{t+1} = \frac{\sigma_{t+1}\hat{x}_{t}+\left(\sigma_{t}-\sigma_{t+1}\right)c_{out}^{t+1}u_{\theta}\left((c_{in}^{t+1}\hat{x}_{t+1}; c_{noise}^{t+1}\right)}{\left(\sigma_{t}-\sigma_{t+1}\right)\left(1-c_{skip}^{t+1}\right)+\sigma_{t+1}}.
\end{equation}
Following previous work  \citep{fan2024videoshop, hertz2022prompt}, during the noise inversion process, we have utilized the noise prediction results at $x_t$ to approximate those at $x_{t+1}$. 

\subsection{Experimental Setting}

\subsubsection{Datasets}
We randomly sample 30 objects from the OmniObject3D Dataset  \citep{wu2023omniobject3d} and render them using Blender's Eevee engine. Each object is rendered in a video sequence comprising 84 frames, with the camera rotating 360/84 degrees between each frame. Additionally, we set the ambient lighting to a white background to match the conditions stipulated by SV3D. It is important to note that, as SV3D has not disclosed the rendering details of its test dataset, achieving pixel-level similarity was challenging.

\subsubsection{The Hyperparameters}
We set the inference steps to 50 with a cfg coefficient of 2.5, following SV3D's configuration, and utilize the Euler sampling method for denoising. Noise optimization comprises 20 steps using a gradient descent optimizer with a learning rate of 1500 and a momentum of 0.5. 

\subsection{Performance Evaluation}

We mainly use Perceptual Similarity (LPIPS  \citep{zhang2018unreasonable}), Structural SIMilarity (SSIM  \citep{wang2004image}), and CLIP similarity score (CLIP-S  \citep{gabriel2021openclip}) to measure the quality of generated results. Due to the lack of multi-view ground truth, pixel-level evaluation metric, such as PSNR, is not applicable. 

The quantitative results in Table \ref{table:noise_optimize_sv3d} demonstrate that optimized noises lead to higher image-to-3D generation quality. 

To facilitate a more intuitive comparison, we also present the qualitative results of original noises and optimized noises in Figure \ref{fig:case_study_3d_0}. It illustrates the significant difference between the optimized noise and the original noise. We can observe that the 3D objects of optimized generally exhibit fewer jagged edges, smoother surfaces, and better fidelity than the results of original noises.

\begin{table}[h]
\caption{The quantitative results of noise optimization for image-to-3D diffusion models according to novel multi-view synthesis on OmniObject3D static orbits.}
\label{table:noise_optimize_sv3d}
\centering
\begin{tabular}{c|c|cc}
\toprule
 Model & Noise & LPIPS$\downarrow$ & SSIM$\uparrow$  \\
\midrule
\multirow{2}{*}{SV3D-U} & Original Noise  & 0.2538  & 0.8664 \\
  & Opti. Noise  & \textbf{0.2523}  & \textbf{0.8768} \\
\bottomrule
\end{tabular}
\end{table}

\section*{Broader Impact} 
\label{sec:impact}

This paper aims at improving diffusion models from the noise-space perspective. While it may have many potential societal consequences, we think none of them must be specifically discussed here.

\end{document}